\documentclass[lettersize,journal]{IEEEtran}
\usepackage{amsmath,amsfonts}
\usepackage{algorithmic}
\usepackage{algorithm}
\usepackage{array}
\usepackage[hidelinks]{hyperref}
\usepackage[caption=false,font=normalsize,labelfont=sf,textfont=sf]{subfig}
\usepackage{textcomp}
\usepackage{stfloats}
\usepackage{url}
\usepackage{verbatim}
\usepackage{graphicx}
\usepackage{tabularx}
\usepackage{cite}
 \usepackage{natbib}
\usepackage{microtype}      
\usepackage{xcolor}         
 \usepackage{amsmath}
 \usepackage{booktabs}
 \usepackage{multirow}
 \usepackage{pifont}
 \usepackage{capt-of}
\usepackage[table]{xcolor}
\usepackage[dvipsnames]{xcolor}

\definecolor{myhighlight}{gray}{0.9}
\usepackage[table]{xcolor}
\begin{document}

\title{
DecoyFace: Beyond Obfuscation via Controllable and Imperceptible Identity Misdirection for Privacy-Preserving Face Recognition
}

\author{Zhihan Ren, Lijun He, Xinyao Wang, Xinzhu Fu, Fan Li,~\IEEEmembership{Senior~Member,~IEEE}
\thanks{Zhihan Ren, Lijun He, Xinyao Wang, Xinzhu Fu, and Fan Li are with
Shaanxi Key Laboratory of Deep Space Exploration Intelligent
Information Technology, School of Information and Communications
Engineering, Xi'an Jiaotong University, Xi'an 710049, China
(e-mail: renzh@stu.xjtu.edu.cn, lijunhe@mail.xjtu.edu.cn, wangxy04@stu.xjtu.edu.cn, xinzhufu@stu.xjtu.edu.cn, lifan@mail.xjtu.edu.cn).}
\thanks{This work is supported by the National Natural Science Foundation of China under Grant No. 62531012, the Sichuan Science and Technology Program under Grant 2026YFHZ0205, and the XJTU Research Fund for AI Science, No.2025YXYC004. \emph{(Corresponding author: Lijun He.)}}
}
\markboth{
Journal of \LaTeX\ Class Files,~Vol.~14, No.~8, August~2021}%
{Shell \MakeLowercase{\textit{et al.}}: A Sample Article Using IEEEtran.cls for IEEE Journals
}

\maketitle

\begin{abstract}
Split face recognition reduces client-side computation but exposes intermediate features to feature inversion attacks and unauthorized analysis by honest-but-curious (HBC) servers.
Existing privacy-preserving face recognition methods mainly aim to resist unauthorized reconstruction, typically producing features whose inversion yields visibly degraded results, which may reveal the existence of protection and motivate adaptive attacks. 
To address this issue, we propose DecoyFace, an imperceptible decoy-oriented framework that steers unauthorized reconstruction toward a plausible but incorrect identity while preserving recognition utility. 
The key idea is to decompose the intermediate representation into a reconstruction-sensitive subspace and its complementary subspace. 
The client injects decoy identity cues into the reconstruction-sensitive subspace, while limited recognition-relevant evidence from the true sample is retained in the complementary subspace.
On the server side, an authorized canonicalization module suppresses decoy-dominant components and recovers a recognition-friendly representation.
This design addresses both attacker-side inversion from intercepted features and HBC server-side reconstruction from canonicalized representations.
Experiments show that DecoyFace preserves competitive recognition accuracy while substantially reducing identity leakage to 2.93\% under U-Net attacks and 0.74\% under Flow-Matching attacks while yielding visually plausible and imperceptible reconstructions, with over 99.78\% face validity on LFW dataset.
\end{abstract}


\section{Introduction}
With the rapid development of deep learning (DL), face recognition (FR) systems have been widely deployed in applications such as mobile authentication, smart access control, and edge-assisted identity verification.
In many real-world scenarios, edge devices have limited computational and memory resources, making it difficult to run a full FR model locally. 
A common solution is to adopt a client-server architecture that offloads compute-intensive stages to a cloud server \cite{kang2017neurosurgeon,teerapittayanon2017distributed}.
However, the transmission of intermediate representations in edge-cloud face recognition raises privacy concerns at both the communication and server computation stages.
(1) On the communication side, a malicious attacker may eavesdrop on the intermediate feature transmitted from the edge device to the cloud server. The attacker then attempts feature inversion from the intercepted representation \cite{singh2024simba,li2023gan,chen2024dia,zhang2024unlocking,lei2025drag,ren2025your}.
(2) On the cloud server side, the honest-but-curious (HBC) server follows the recognition protocol but may additionally inspect the received feature or recovered canonicalized representations for unauthorized reconstruction. 
Therefore, the desired defense should not only preserve recognition accuracy for authorized verification but also prevent both intercepted and server-observed representations from revealing the original identity. 
\begin{figure}
    \centering
    \includegraphics[width=\linewidth]{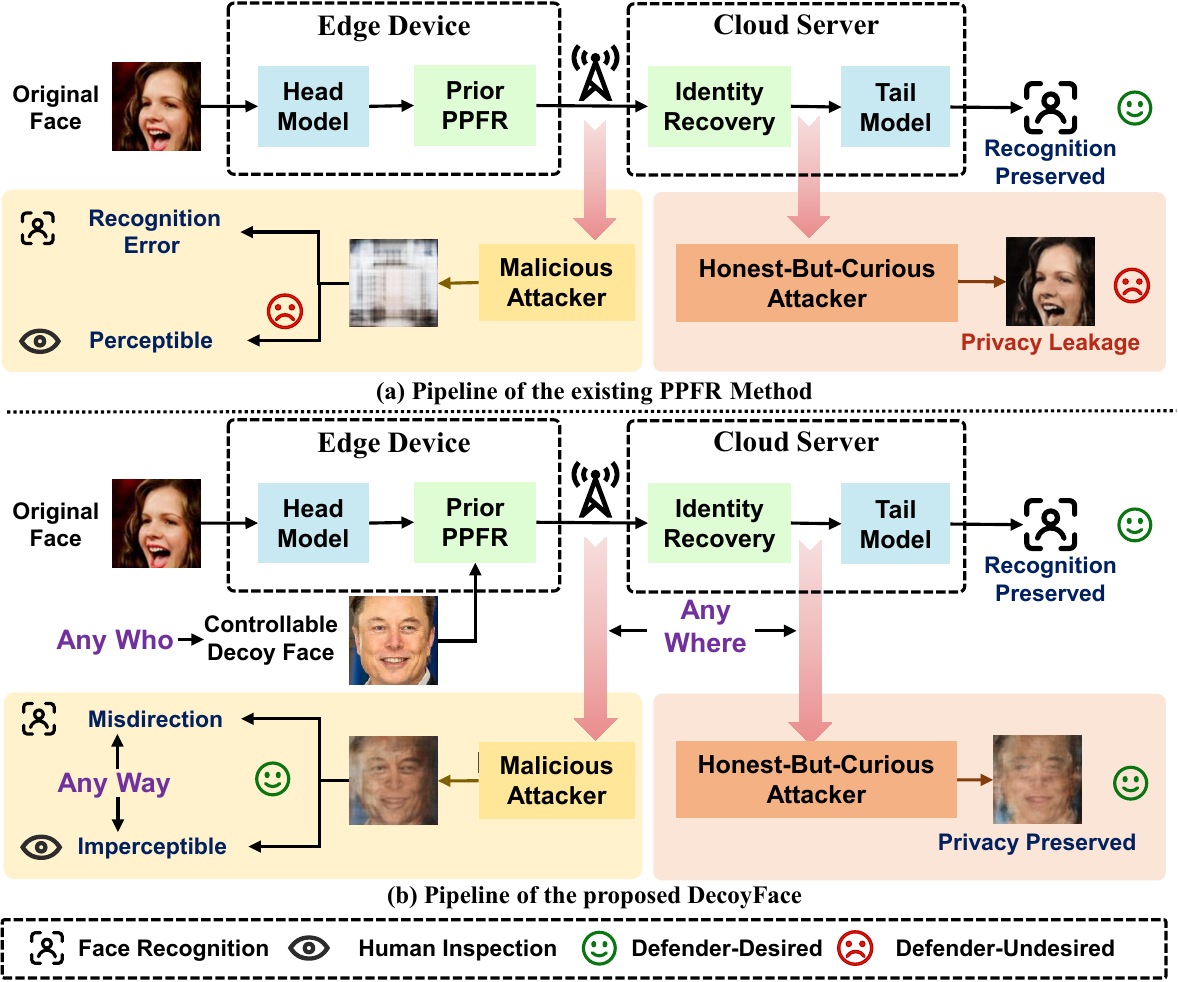}
    \caption{Comparison between existing PPFR methods (Top Row) and the proposed DecoyFace framework (Bottom Row). 
    Existing PPFR methods primarily obfuscate transmitted features, producing distorted reconstructions that reveal the presence of protection.
    In contrast, DecoyFace steers unauthorized inversion toward a realistic but misleading decoy identity\protect\footnotemark[1]. 
    }
    \label{fig:placeholder}
\end{figure}

\footnotetext[1]{ The decoy face shown for illustration is cropped from \href{https://commons.wikimedia.org/wiki/File\%3AElon_Musk_close-up_\%28cropped\%29.jpg}{this URL.}
}
\setcounter{footnote}{1}

As summarized in Table \ref{tab:table1}, prior studies can be categorized according to their operating domain and privacy objective.
Image-domain approaches, such as deepfake defense methods \cite{li2025anti,wang2025faceswapguard,wang2025nullswap}, perturb the input image to disrupt downstream misuse, while virtual-ID methods \cite{yuan2022generating,wang2024make} directly synthesize a substitute identity. 
Nevertheless, our objective differs from virtual-ID generation methods. 
Rather than replacing the user identity with a new enrolled identity for downstream recognition, we preserve recognition for the true identity on the authorized path while misleading unauthorized inference on the leaked representation.
In the frequency and feature domains, feature encryption \cite{hsu2012image,ao2025cryptoface} seeks to make transmitted representations hard to exploit, and privacy-preserving face recognition (PPFR) methods \cite{ji2022privacy,wang2023privacy,mi2022duetface,mi2024privacy,jin2024faceobfuscator,daifracface} mainly learn inversion-resistant features that preserve recognition utility while degrading visual recovery.
However, many obfuscation-oriented PPFR methods treat failed or visibly distorted reconstructions as successful privacy outcomes.
In summary, existing methods face two limitations: \textbf{(1) Limited protection coverage:} They mainly protect against feature leakage during wireless transmission, while overlooking unauthorized reconstruction performed by HBC servers from recovered representations; \textbf{(2)} \textbf{Lacking Concealed Defense Capabilities}: Visibly distorted reconstructions may reveal the presence of protection, thereby encouraging attackers to refine their inversion models or adopt stronger attacks. 
It is necessary to redirect unauthorized reconstruction toward a plausible but incorrect identity, making the privacy protection less perceptible and harder to adapt against.

\begin{table*}[t]
\centering
\setlength{\tabcolsep}{3pt}
\renewcommand{\arraystretch}{0.9}
\caption{Comparison of different privacy protection directions for FR. \(A\) denotes the original identity, while \(B\) denotes the decoy identity. 
}
\label{tab:comparison_misdirection}
\begin{tabular}{lllll}
 \toprule
 \textbf{Category} & \textbf{Domain} & \textbf{Privacy objective}  &\textbf{Authorized utility}& \textbf{Attacker outcome} \\
 \midrule
 No protection & None & None  &Preserved&Original ID \(A\) \\
 Deepfake defense \cite{li2025anti,wang2025faceswapguard,wang2025nullswap}& Image & Perturbed image  &N/A& Not reliably \(A\) \\
 Virtual ID \cite{yuan2022generating,wang2024make} & Image & Generate virtual ID  &Preserved for virtual ID & Virtual ID \\
  Feature encryption \cite{hsu2012image,ao2025cryptoface}& Feature & Encrypted features  &Preserved & Hard to exploit \\
 PPFR \cite{wang2023privacy,mi2022duetface,jin2024faceobfuscator,daifracface}& Frequency / Feature & Inversion-resistant features  &Preserved& Distorted recovery \\
 \textbf{Ours} & \textbf{Feature} & \textbf{Misdirection to decoy ID \(B\)}  &\textbf{Preserved}& \textbf{Reconstructed as decoy ID \(B\)} \\
    \bottomrule
  \end{tabular}
\label{tab:table1}
\end{table*}

To address these limitations, we propose DecoyFace, an imperceptible decoy-oriented privacy-preserving framework for face recognition.
Unlike existing obfuscation-oriented PPFR methods, DecoyFace preserves authorized recognition utility while misleading unauthorized reconstruction toward a plausible but incorrect decoy identity. 
This shift from reconstruction failure to identity misdirection distinguishes our method from existing privacy-preserving approaches and provides a new perspective for secure feature transmission in edge-cloud FR.
More importantly, we consider not only attacker-side feature inversion from intercepted features, but also unauthorized reconstruction by an HBC server from the server-side canonicalization representation. 
This broader threat model better reflects the privacy risks of practical split face recognition systems.
Fig. 1 illustrates the conceptual difference between existing obfuscation-oriented PPFR methods and the proposed decoy-based misdirection framework.
The key idea is to exploit the asymmetry between unauthorized reconstruction and authorized recognition in the intermediate feature space. 
Specifically, we first decompose the transmitted feature into a reconstruction-sensitive subspace and its complementary subspace via Reconstruction-Sensitive Subspace Decomposition (RSSD). 
The former captures the feature directions that are most influential for attacker-side inversion, while the latter serves as a carrier of recognition-relevant evidence in a form less susceptible to reconstruction. 
Based on this decomposition, we design a client-side Decoy-guided Coherent Mixing (DGCM) that injects decoy identity cues into the sensitive component, and a server-side Authorized Canonicalization Module (ACM) that suppresses these cues for authorized verification. 
By treating leaked features as opportunities for controlled deception, DecoyFace induces attackers to reconstruct a realistic yet incorrect identity.

The main contributions are summarized as follows:
\begin{enumerate}
\item  \textbf{Any Where: leakage protection across wireless transmission and HBC servers.}
Unlike existing privacy-preserving FR methods that mainly defend against feature leakage during wireless transmission, DecoyFace jointly protects both the communication side and the HBC server side, where unauthorized reconstruction may be performed from recovered representations.
\item  \textbf{Any Way: imperceptible resistance to both machine-vision and human inspection.}
We develop a reconstruction-sensitive privacy protection mechanism, which decomposes the feature into reconstruction-sensitive and insensitive subspaces, injects decoy cues into the sensitive component, and enables authorized recognition while avoiding perceptible distortions that may expose the presence of protection.
\item  \textbf{Any Who: identity-controllable misdirection.}
We introduce a decoy-oriented privacy objective that steers unauthorized reconstruction toward a plausible but incorrect identity, instead of degrading reconstruction quality. 
This design reduces true-identity leakage while preserving facial plausibility in unauthorized outputs, thus enabling stealthy identity misdirection.
\end{enumerate}

\section{Related Work}
\label{sec:related}
\subsection{Face Reconstruction Attacks}
Face reconstruction attacks aim to recover private visual content from transmitted or stored representations. 
Recent studies have shown that intermediate features in split inference can leak rich visual and identity-related information, making them vulnerable to feature inversion. 
Singh~et~al.~\cite{singh2024simba} benchmarked split inference under different split points and adversarial knowledge settings, showing that leaked features can be reconstructed with nontrivial fidelity. 
Benefiting from the growth of generative models in image synthesis and reconstruction \cite{song2019generative,ren2023context,he2024unsupervised,ren2024super,wang2025beyond,bai2026blind}, subsequent work has strengthened attackers by introducing stronger generative priors.
GLASS \cite{li2023gan} proposed GAN-based attacks that search in a pretrained latent space to improve the realism and stability of recovered images. 
More recent approaches further exploit diffusion \cite{song2019generative,song2020score} or flow-based \cite{lipman2022flow} priors. 
DRAG \cite{lei2025drag} used guided diffusion to reconstruct higher-fidelity images from intermediate representations.
FIA-Flow \cite{ren2025your} studied data-efficient black-box inversion and showed that semantically faithful recovery is possible even without access to the victim model’s internal parameters. 
These results indicate that intermediate features are sufficient to recover identifiable faces in split face recognition.

\subsection{Privacy-Preserving Face Recognition}
Existing PPFR methods can be broadly grouped into three lines. 
The first line relies on cryptographic protection, such as homomorphic-encryption-based matching or end-to-end encrypted face recognition, which offers strong confidentiality guarantees but often introduces substantial computational overhead and deployment complexity \cite{hsu2012image,ao2025cryptoface}. 
The second line operates in the image domain. 
Deepfake defense \cite{li2025anti,wang2025faceswapguard,wang2025nullswap} and identity-cloaking methods perturb the original face image to prevent malicious reuse, while virtual-ID or cancelable-face methods generate substitute identities for downstream recognition. 
These methods are valuable for image sharing and media protection, but they do not directly address privacy leakage from intermediate features during split inference. 
Moreover, virtual-ID methods \cite{yuan2022generating,wang2024make} usually preserve utility for the substitute identity rather than the true enrolled identity.
The third and most relevant line is representation-domain PPFR methods, which attempt to preserve recognition utility while suppressing unauthorized reconstruction from transmitted features. 
AdvFace \cite{wang2023privacy} learns privacy-preserving adversarial facial features as a plug-in defense against unknown reconstruction models. 
MinusFace \cite{mi2024privacy} removes reconstruction-friendly cues via trainable feature subtraction and randomized transformation. 
FaceObfuscator \cite{jin2024faceobfuscator} further constructs gradient-descent-resistant obfuscated features, while FracFace \cite{daifracface} disrupts spatial regularities in the frequency domain to weaken visual cues exploitable by generative reconstruction.

However, most existing PPFR methods aim to make unauthorized reconstruction fail or appear visually distorted.
By contrast, DecoyFace adopts a misdirection-oriented design that aims to preserve verification utility for the server while steering unauthorized reconstruction toward a plausible but incorrect identity. 
\section{Preliminary}
\begin{figure*}
    \centering
    \includegraphics[width=0.95\linewidth]{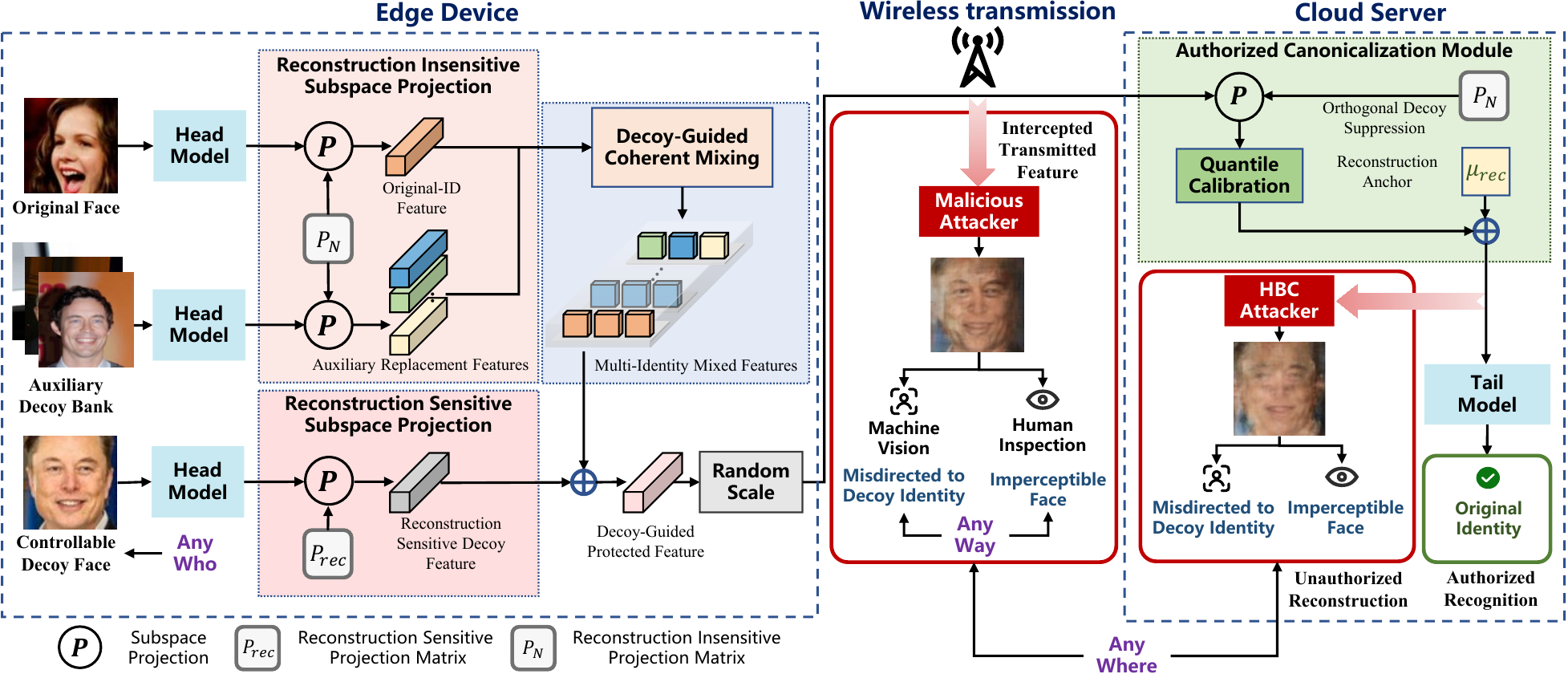}
    \caption{Overview of the proposed DecoyFace framework. The edge device decomposes the intermediate representation, injects decoy cues into the reconstruction-sensitive component, and constructs a protected complementary carrier via DGCM. The cloud server then suppresses the decoy-dominant component and canonicalizes the received feature for authorized face recognition.}
    \label{fig:pipeline}
\end{figure*}
\subsection{Threat Model}
\subsubsection{Attacker's Scenarios and Knowledge}
In this work, we consider a threat model involving both a \textbf{malicious attacker} and an \textbf{honest-but-curious server}.

The malicious attacker can intercept transmitted features during edge-cloud communication. 
The client encoder $F_c$ maps a face image $x$ to an intermediate feature $z = F_c(x)$ and transmits $z$ to a cloud server $F_s$ for identity verification.
The attacker is assumed to have black-box query access to the client model. 
This assumption follows recent feature inversion studies and reflects practical queryable edge clients \cite{ren2025your,zhang2024unlocking,chen2024dia}.
It can submit inputs $x_i$ and obtain the corresponding output features $z_i$, but cannot access the model architecture or parameters.

In addition, we consider an HBC server that follows the normal recognition protocol but may additionally analyze the received protected features and the server-side recovered representations for unauthorized privacy inference. 
This models a stronger and more realistic privacy risk in deployed split FR systems, where the server is trusted for utility but not necessarily for privacy.
\subsubsection{Attacker's Strategy}
Using queried image-feature pairs $\{x_i,z_i\}$, the attacker trains a reconstruction model to reconstruct face images from intercepted features.
In particular, the HBC server may apply the server-side projection and canonicalization used for authorized recognition and then attempt reconstruction from that de-randomized representation.

\subsection{Defender Objective}
Existing PPFR methods mainly aim to transmit a protected feature $\tilde{z}$ that preserves recognition utility while reducing reconstruction fidelity, so that reconstructions from $\tilde{z}$ become visually distorted.
In contrast, our goal is to reduce true-identity leakage while maintaining the facial plausibility of unauthorized outputs.
This objective is more targeted than merely degrading reconstruction quality, because it steers unauthorized reconstructions toward a plausible yet incorrect identity rather than merely revealing that protection has been applied.  

Accordingly, the defender should satisfy the following three requirements simultaneously:
\begin{enumerate}
\item \textbf{Authorized utility}: The protected feature should remain discriminative for the true identity under the authorized recognition pipeline.

\item \textbf{Attacker-side privacy}: The protected feature should reduce similarity to the original image and suppress identity leakage while maintaining facial plausibility in unauthorized reconstruction.

\item \textbf{Server-side privacy}: Even after server-side authorized processing, the HBC server should not enable faithful reconstruction of the original face from the recovered feature.
\end{enumerate}

\section{Method}
\subsection{Overview of the Proposed Framework}
Given a face image $x_A$ of identity $\mathcal A$, the client encoder extracts an intermediate feature $z_A = F_c(x_A)$. 
DecoyFace transforms $z_A$ into a protected representation $\tilde{z}$ that preserves utility for authorized verification while biasing unauthorized reconstruction toward a decoy identity.

As shown in Fig.~\ref{fig:pipeline}, Reconstruction-Sensitive Subspace Decomposition (RSSD) separates the feature space into a reconstruction-sensitive subspace and a complementary subspace.
Based on this decomposition, Decoy-Guided Coherent Mixing (DGCM) injects decoy identity cues into the sensitive subspace while preserving limited true-identity evidence in the complementary component.
Finally, the Authorized Canonicalization Module (ACM) provides a constrained recovery path that supports authorized verification without fully restoring the original protected feature. 
Overall, DecoyFace redirects leakage by injecting decoy cues into reconstruction-sensitive directions while retaining limited recognition-relevant evidence in the complementary subspace. 
The resulting representation is easier for the authorized server to canonicalize than for an attacker to exploit directly. 
This asymmetry is the basis of the proposed client-server PPFR design.

\begin{figure*}
    \centering
    \includegraphics[width=\linewidth]{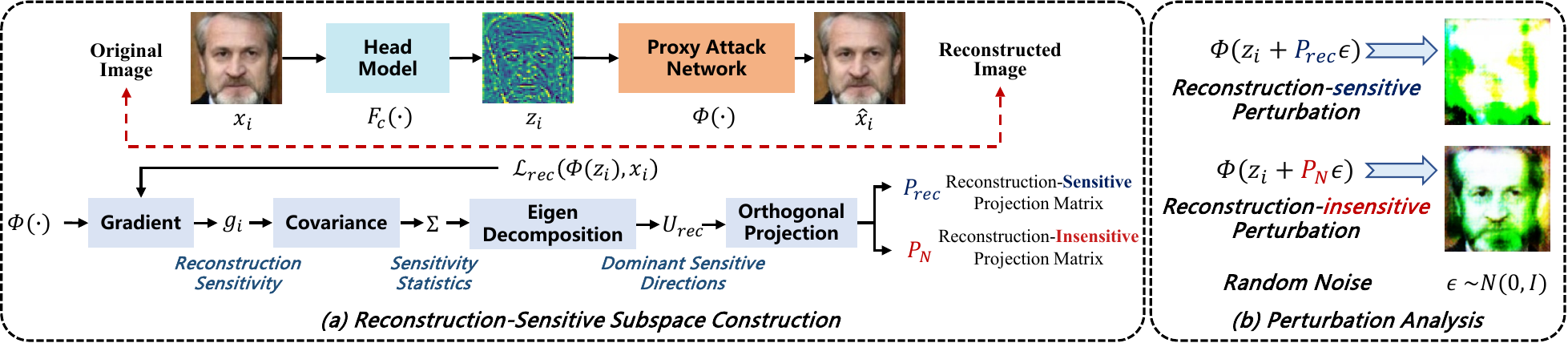}
    \caption{Construction and illustration of the reconstruction-sensitive subspace.
    Perturbations restricted to the RSS induce substantially larger reconstruction changes than perturbations of equal magnitude applied to the complementary subspace.}
    \label{fig:rssd}
\end{figure*}
\subsection{Reconstruction-Sensitive Subspace Decomposition}
\label{sec:subspace}
The transmitted feature serves two competing purposes: it must preserve identity information for recognition, yet it also exposes information that can be exploited for unauthorized reconstruction.
Direct perturbation does not distinguish directions that primarily support recognition from those that primarily support reconstruction, and therefore makes it difficult to inject decoy cues in a controlled manner.
To address this problem, we explicitly decompose the intermediate feature space into a reconstruction-sensitive subspace (RSS) and a complementary subspace, enabling the two roles to be manipulated separately.

As shown in Fig. \ref{fig:rssd}, we estimate the subspace using an offline proxy inversion network $\Phi(\cdot)$, whose architecture need not match the attacker used in the evaluation.
For each input image $x_i$, the client encoder produces the intermediate feature $z_i = F_c(x_i) \in \mathbb{R}^{C\times H\times W}$,
and the proxy attacker reconstructs $ \hat{x}_i = \Phi(z_i)$. 
We define the reconstruction loss as $\mathcal{L}_{\mathrm{rec}} = \ell_{1}(\hat{x}_i, x_i).$
To measure reconstruction sensitivity with respect to the intermediate feature, we compute the gradient of the reconstruction loss $g_i =\nabla_{z_i}\mathcal{L}_{\mathrm{rec}}$.

\begin{figure*}
    \centering
    \includegraphics[width=0.95\linewidth]{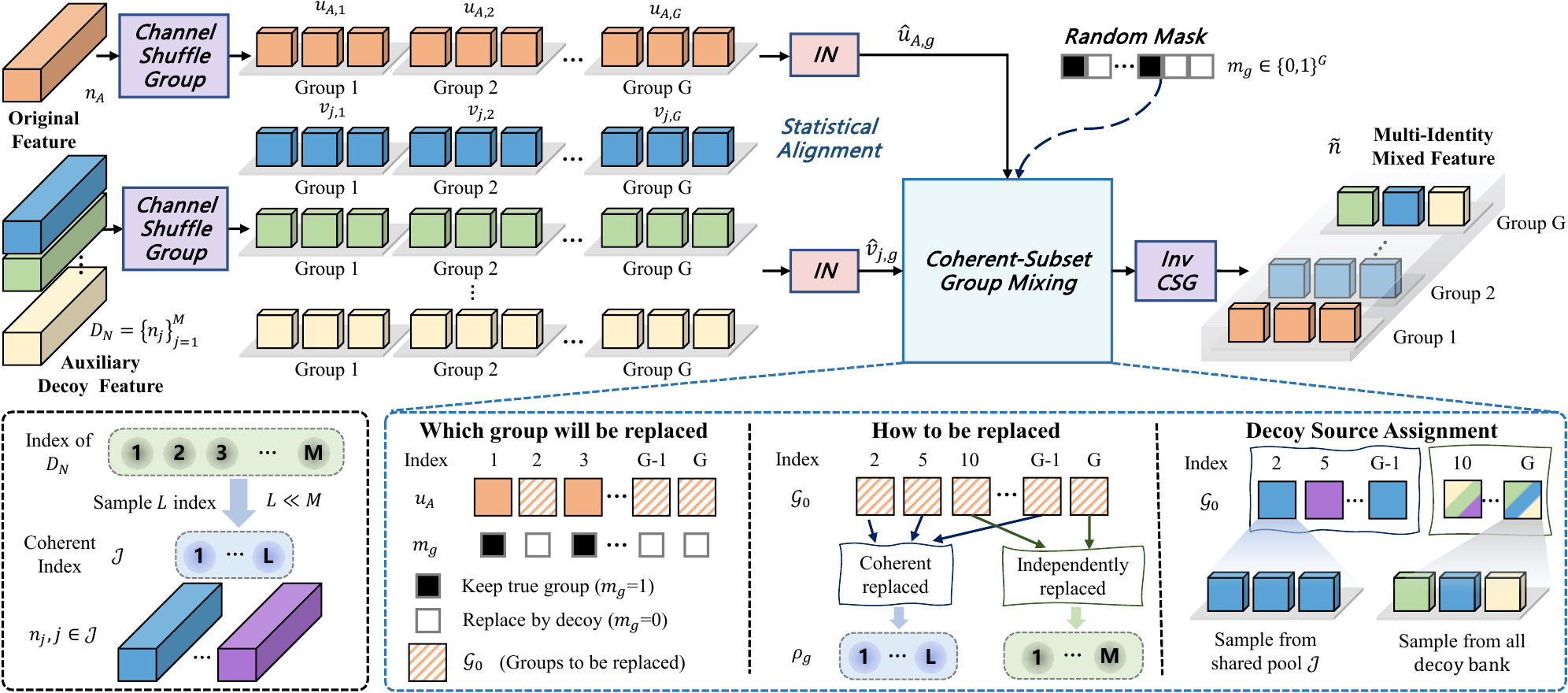}
    \caption{Detailed design of Decoy-Guided Coherent Mixing (DGCM). DGCM first applies channel-shuffled grouping (CSG) to partition complementary features into groups, followed by Instance Norm (IN) to normalize group-wise statistics. A routing mask preserves a small subset of true groups, while the remaining groups are replaced by coherent or independently sampled decoy groups. Inv-CSG maps the mixed groups back to the original feature layout to form the protected complementary carrier.}
    \label{dgcm}
\end{figure*}
Intuitively, feature directions whose perturbations consistently induce large reconstruction changes across samples are more likely to encode recoverable facial structure. 
Therefore, we identify the RSS from the dominant eigensubspace of the gradient covariance.
\begin{align}
\bar g_i[c] =
\frac{1}{HW}\sum_{h=1}^{H}\sum_{w=1}^{W}
g_i[c,h,w], \quad
\mu_g =
\frac{1}{N}\sum_{i=1}^{N}\bar g_i, \\ 
\Sigma = \frac{1}{N-1}\sum_{i=1}^{N}(\bar g_i-\mu_g)(\bar g_i-\mu_g)^\top\in \mathbb{R}^{C\times C}.
\end{align}
The dominant eigenvectors correspond to feature perturbations that consistently produce large reconstruction variations across samples, and are treated as reconstruction-sensitive.
We then perform eigendecomposition of $\Sigma = U \Lambda U^\top$. 
Let $U_{\mathrm{rec}}$ denote the top-$r$ eigenvectors.
These vectors define the dominant RSS, and the corresponding projection matrix is $P_{\mathrm{rec}} = U_{\mathrm{rec}} U_{\mathrm{rec}}^\top \in \mathbb{R}^{C \times C}$.
Its complementary projection is defined as $P_{\mathrm{N}} = I - P_{\mathrm{rec}}$.
Accordingly, any intermediate feature $z$ can be decomposed as
\begin{equation}
z = P_{\mathrm{rec}} z + P_{\mathrm{N}} z,
\end{equation}
where $P_{\mathrm{rec}} z$ denotes the component that is most influential for reconstruction.
 
As shown in Fig. \ref{fig:rssd}, perturbations restricted to the RSS produce substantially larger reconstruction changes than perturbations of the same magnitude applied to the complementary subspace. 
By contrast, perturbations of comparable magnitude in the insensitive subspace, $\Phi(z_i + P_{\mathrm{N}} \epsilon)$, have a much weaker visual effect. 
This observation motivates using $P_{\mathrm{rec}}$ as the primary subspace for decoy injection, and treating $P_{\mathrm{N}}$ as the complementary carrier for preserving recognition-relevant information on the authorized path.
It also enables a matched recovery rule on the server side, where the decoy-dominant component can be suppressed through projection and subsequent canonicalization. 
This decomposition provides the structural basis for the subsequent asymmetric design: decoy cues are injected into the RSS, whereas the complementary subspace is used as a constrained carrier for recognition.

\subsection{Client-side Protected Feature Generation}
\label{sec:DGM}
Given the RSSD, the next question is how to construct a protected feature that remains discriminative for the authorized server while redirecting unauthorized inversion toward a decoy identity.
Replacing the sensitive component is insufficient because the complementary branch would remain unchanged. 
An unchanged complementary branch may still preserve excessive true-identity information and may also create statistical inconsistency between the two branches, thereby weakening controllability under both inversion and server-side recovery.
Therefore, we design a structured client-side mixing process that retains only a limited amount of true-identity evidence while replacing the remaining complementary groups with auxiliary complementary components from decoy samples. 
To reduce fragmentation among independently replaced groups, we further design a coherent-subset group mixing policy, under which a subset of the replaced groups shares a small number of auxiliary sources.

Let $x_A$ be an input image from the true identity $\mathcal{A}$, and let $y_B$ be a decoy sample from another identity $\mathcal{B}$. 
Their encoder features are $z_A = F_c(x_A), \quad z_B = F_c(y_B).$
We decompose them into reconstruction-sensitive and insensitive components:
\begin{equation}
r_B = P_{\mathrm{rec}} z_B, \quad n_A = P_{\mathrm{N}} z_A, 
\end{equation}
where $r_B$ denotes the reconstruction-sensitive component extracted from identity $\mathcal{B}$ and $n_A$ denotes the complementary component of the true sample $\mathcal{A}$.
Meanwhile, we maintain an auxiliary decoy sample set $y_{\mathrm{aux}}=\{y_j\}_{j=1}^{M}$, which is not tied to the current target decoy identity $B$.
The corresponding auxiliary decoy feature bank $\mathcal D_N=\{n_j\}_{j=1}^M$ is defined as 
\begin{equation}
n_j=P_{\mathrm{N}}F_c(y_j), \quad n_j\in\mathbb R^{C\times H\times W},
\end{equation}
where each $n_j$ is extracted from an auxiliary sample $y_j$ and projected onto the reconstruction-insensitive subspace. 
The auxiliary components serve both to fill the suppressed complementary groups and to reduce the risk that the server reconstructs a clean complementary branch associated with the true identity.

The detailed design of DGCM is illustrated in Fig. 4.
We operate on the complementary branch at the group level. 
The key idea is to retain only a limited subset of complementary groups from the true sample and to replace the remaining groups with auxiliary complementary groups drawn from $\mathcal D_N$.
This design controls the amount of true-sample evidence exposed to the server while making the transmitted complementary branch less cleanly attributable to the true sample.
Specifically, a channel-shuffled grouping operator (CSG) $\Gamma(\cdot)$ partitions the feature into $G$ groups: \begin{equation}
\begin{aligned}
\Gamma:\mathbb R&^{C\times H\times W}\rightarrow \prod_{g=1}^{G} \mathbb R^{c_g\times H\times W},\sum_{g=1}^{G}c_g=C, \\
u_A &= \Gamma(n_A)=(u_{A,1},\ldots,u_{A,G}),\\
v_j &= \Gamma(n_j)=(v_{j,1},\ldots,v_{j,G}).
\end{aligned}
\end{equation}

To control how much true-sample evidence is retained, we randomly select $K$ complementary groups from $n_A$, yielding a binary routing mask $m \in\{0,1\}^G$  with $\|m\|_0=K$.
Groups with $m_g = 1$ are preserved from $n_A$, while the remaining suppressed groups with $m_g = 0$ are replaced by auxiliary groups from $\mathcal D_N$. 
A straightforward replacement strategy is to sample the source of each suppressed group independently. 
However, fully independent replacement tends to produce a fragmented complementary branch, in which replaced groups originate from unrelated decoy samples and thus exhibit inconsistent cross-group statistics, which may be exploited by the attacker. 

To alleviate this issue, we introduce a coherent-subset group mixing policy that divides the replaced groups into coherent and independently replaced subsets. 
Let $\mathcal G_0=\{g\in\{1,\ldots,G\}:m_g=0\}$ denote the set of suppressed groups. 
We uniformly sample a coherent subset
\begin{equation}
    \mathcal S\sim \mathrm{Unif}\left(\{S\subseteq \mathcal G_0:|S|=s_c\}\right),
\end{equation}
where $s_c\le |\mathcal G_0|$. 
The remaining groups $\mathcal G_0\setminus\mathcal S$ are independently replaced.
We further sample a small shared source index set
\begin{equation}
    \mathcal J\sim \mathrm{Unif}\left(\{J\subseteq \{1,\ldots,M\}:|J|=L\}\right),
\end{equation}
where $L\le s_c$ and $L \ll M$. Here, $\mathcal S$ specifies which suppressed groups use coherent replacement, while $\mathcal J$ provides the shared pool of auxiliary source indices available to these groups.
Conditioned on $\mathcal S$ and $\mathcal J$, the source index for each suppressed group is sampled as
\begin{equation}
\rho_g\sim\begin{cases}\mathrm{Unif}(\mathcal J), & g\in\mathcal S,\\\mathrm{Unif}(\{1,\ldots,M\}), & g\in\mathcal G_0\setminus\mathcal S.\end{cases}
\end{equation}

In this way, only a subset of the replaced groups exhibits controlled source sharing. This encourages higher cross-group source consistency in the replaced branch, while keeping the number of preserved true-sample groups unchanged.
Before mixing, we normalize each group independently to reduce source-dependent magnitude mismatch:
\begin{equation}
\hat{u}_{A,g} =\mathrm{IN}\!\left(u_{A,g}\right), 
\hat{v}_{j,g} =\mathrm{IN}\!\left(v_{j,g}\right), 
\end{equation}
where $\mathrm{IN}(\cdot)$ denotes the InstanceNorm \cite{ulyanov2016instance} operator.
The mixed complementary feature is constructed group-wise as
\begin{equation}
    \tilde{u}_g =\begin{cases}\hat{u}_{A,g}, & m_g = 1, \\\hat{v}_{\rho_g,g}, & m_g = 0,\end{cases}
\end{equation}

Finally, we apply the inverse grouping operator to recover the mixed complementary component:
\begin{equation}
\tilde n = \Gamma^{-1}(\tilde u_1,\ldots,\tilde u_G).
\end{equation}
This design improves local consistency in the mixed complementary branch without increasing the amount of retained true-sample evidence. 
Specifically, the amount of retained true-sample information is still determined by the $K$ routed true groups selected by $m$. The coherent-subset group mixing policy only reorganizes the replaced groups and never introduces additional information from $n_A$. The complementary branch is not a clean residual of the true sample, but a constrained carrier that preserves only limited true-identity evidence while exhibiting more stable cross-group statistics than fully independent replacement.
 
The final protected feature is constructed as
\begin{equation}
\tilde{z} = \tilde{n} + r_B.
\end{equation}
The reconstruction-sensitive branch is dominated by decoy identity cues, whereas the complementary branch contains only sparse routed evidence from the true sample, together with auxiliary complementary content from other samples.
Importantly, the main decoy feature $r_B$ is sampled once per protected view. 
In contrast, the complementary replacement groups are drawn from the auxiliary bank and are not tied to the identity of $\mathcal{B}$.
Even if the server can suppress the decoy, the remaining complementary branch does not coincide with the original $n_A = P_{\mathrm{N}} z_A$.

Before transmission, we apply a sample-wise random global scaling transform to the protected feature $\tilde z$.
Specifically, we sample an integer exponent $k\in \{ -t,-t+1,\ldots,t-1,t \}$, set $s=2^k$, and transmit $z_t=s\tilde z$. The scaling factor is not revealed to the server and is instead handled implicitly by server-side canonicalization.
The server-side ACM is designed to absorb this magnitude variation through matched canonicalization.

Consequently, the transmitted feature is no longer a simple perturbation of the true sample, but a structured mixture whose leakage is biased toward the decoy identity and whose complementary branch is intentionally constructed to be only partially recoverable on the authorized server side.

\subsection{Server-side Recognition}
\label{sec:server}
After receiving the protected feature $z_t$, the server performs a matched recovery procedure that suppresses the reconstruction-sensitive decoy component and restores a representation more suitable for downstream recognition.
Since the decoy cue is injected primarily through the RSS, the server first suppresses this decoy-dominant component by projecting the received feature onto the reconstruction-insensitive subspace:
\begin{align}
z_s  &= P_{\mathrm{N}} z_t=s P_{\mathrm{N}} (\tilde{n} + r_B)  \\ \nonumber
&=s P_{\mathrm{N}} \tilde{n}+s P_{\mathrm{N}} P_{\mathrm{rec}}z_B
=s P_{\mathrm{N}} \tilde{n},
\end{align}
This cancellation holds because $r_B\in \mathrm{range}(P_{\mathrm{rec}})$, and $P_{\mathrm{N}}P_{\mathrm{rec}}=0$.
This projection suppresses the RSS component that contains the explicitly injected decoy cues and retains only the projected complementary carrier, still scaled by the unknown factor $s$.
As a result, $z_s$ is expected to preserve the recognition-relevant evidence from the routed complementary groups, while discarding the decoy-dominant component injected into the RSS.

To reduce the magnitude perturbation introduced during transmission, the server further performs group-wise canonicalization on $z_s$.
We adopt quantile-based calibration instead of normalization because the former is less sensitive to outlier activations introduced by group replacement and feature mixing.
We partition $z_s$ into $G$ groups and rescale each group independently according to its $q$-quantile magnitude:
\begin{equation}
z_{\mathrm{cal}}^{(g)} =
\frac{\tau_s} {Q_q(|z_s^{(g)}|)+\epsilon}\, z_s^{(g)},
\quad g=1,\ldots,G,
\end{equation}
where $z_s^{(g)}$ denotes the $g$-th feature group, $Q_q(\cdot)$ denotes the $q$-quantile computed over the absolute values of all elements in the group, $\tau_s$ is the target server-side magnitude, and $\xi$ is a small constant for numerical stability.
After calibrating all groups, we reassemble them to obtain the calibrated feature $n_{\mathrm{cal}}=\Gamma^{-1}(z^{(1)}_{\mathrm{cal}},\ldots,z^{(G)}_{\mathrm{cal}}) $.

The server then forms a canonicalized recognition feature by combining the calibrated complementary component with a fixed anchor in the RSS:
\begin{equation}
z_{\mathrm{can}} = \mu_{\mathrm{rec}} + n_{\mathrm{cal}},
\end{equation}
where $\mu_{\mathrm{rec}}$ is defined as the mean reconstruction-sensitive component of the decoy bank 
\begin{equation}
\mu_{\mathrm{rec}}=\frac{1}{M}\sum_{j=1}^{M}P_{\mathrm{rec}}F_c(y_j),
\end{equation}
and serves as a fixed reference in the $P_{\mathrm{rec}}$ subspace, thereby reducing distribution shift after the decoy-dominant component has been removed.
The canonicalized feature $z_{\mathrm{can}}$ is finally fed into the server backbone for recognition:
\begin{equation}
e=F_s(z_{\mathrm{can}}),\quad p=H_{\mathrm{arc}}(e),
\end{equation}
where $H_{arc}$ denotes the ArcFace classification head \cite{deng2019arcface}.  
From the attacker's perspective, the transmitted feature is dominated by reconstruction-sensitive decoy cues and thus tends to produce a plausible but incorrect face under inversion. 
In contrast, the authorized server applies the matched projection and scale calibration to recover a more stable recognition feature for downstream authentication.

\subsection{Training Objective}
\label{sec:loss}
The client-side protection process is stochastic because the routing mask, complementary replacement, and transport-time scaling vary across protected views. 
Identity supervision alone is insufficient because it may preserve class separability while still permitting large view-dependent variation across protected samples of the same input.
The server backbone $F_s$ and the ArcFace classification head  $H_{arc}$ are trained jointly on the recovered features. 
During training, we generate $V$ protected views for each sample using different client-side randomization, and recover their corresponding server-side features $\{z^{v}_{\mathrm{can}}\}_{v=1}^{V}$.

The overall training objective is
\begin{equation}
\mathcal{L}
=
\mathcal{L}_{\mathrm{id}}
+
\lambda_{\mathrm{cons}} \mathcal{L}_{\mathrm{cons}}.
\end{equation}

For identity supervision, we apply the ArcFace classification loss on the canonicalized  feature:
\begin{equation}
\mathcal{L}_{\mathrm{id}}=\frac{1}{V}\sum_{v=1}^{V}\mathrm{CE}\bigl(H_{arc}(F_s(z^{v}_{\mathrm{can}})),\, y_i\bigr).
\end{equation}
where $\mathrm{CE}(\cdot)$ denotes the cross-entropy loss, and $y_i$ is the ground-truth identity label of the training image $x_i$.
This term encourages the recovered representation to remain discriminative for the true identity after client-side protection and server-side recovery.

To reduce variation across protected views, we further impose a multi-view embedding consistency loss. Let
\begin{equation}
\hat e^v
=
\frac{F_s(z^{v}_{\mathrm{can}})}{\|F_s(z^{v}_{\mathrm{can}})\|_2}
\end{equation}
denote the normalized embedding of the $v$-th protected view. 

The consistency loss is defined as
\begin{equation}
\mathcal{L}_{\mathrm{cons}}=\frac{2}{V(V-1)}\sum_{1 \le u < v \le V}\left\|\hat{e}^u - \hat{e}^v\right\|_2^2.
\end{equation}
This term stabilizes the server-side embedding against stochastic client-side protection, which is important because the same input may generate multiple protected views with different routed groups, auxiliary assignments, and transport-time scales.
\section{Experiments}
\subsection{Datasets and Metrics}
We use MS1Mv2 \cite{guo2016ms} as the training set and evaluate on LFW \cite{huang2008labeled}, AgeDB \cite{moschoglou2017agedb}, CFP-FP \cite{sengupta2016frontal}, CALFW \cite{calfw}, CPLFW \cite{cplfw}, IJB-B \cite{ijbb}, and IJB-C \cite{ijbc}. 
We report TAR@FAR=$10^{-4}$ on IJB-B/IJB-C and verification accuracy (Acc) on the other datasets.

We evaluate the method on authorized recognition and privacy protection against both malicious attackers and the HBC server.
The experiments follow the edge-cloud split FR setting, where the client transmits intermediate features to the server for identity verification. 
Unlike conventional PPFR evaluations that primarily degrade reconstruction quality, our evaluation also examines whether the reconstructed results are redirected toward a decoy identity while preserving utility for the authorized server.
Cosine similarity (COS) in the identity embedding space is used to assess identity consistency.

In addition, to better reflect the decoy-oriented goal of our method, we use three identity-related metrics to evaluate reconstructed results. 
Let \(x_i\) be the original image with identity label \(y_i\), and let \(\hat{x}_i\) be the reconstructed image. 
We first apply a face detector \(d(\cdot)\) to \(\hat{x}_i\) and use its maximum detection confidence as the face-validity score. Given a threshold \(\theta_f\), the valid reconstruction set and the face validity ratio (FVR) are defined as
\begin{equation}\mathcal{V}=\{i \mid d(\hat{x}_i)>\theta_f\}, \qquad\mathrm{FVR}=\frac{|\mathcal{V}|}{N}.
\end{equation}

For identity evaluation, we use an independent face recognition model \(\phi(\cdot)\), with all embeddings normalized to unit length. Let \(\mathcal{T}\) denote the closed-set identity gallery, and let \(\mathbf{t}_c\) be the identity template of class \(c\in\mathcal{T}\). For each valid reconstruction, the predicted identity is obtained by top-1 nearest-template matching:
\begin{equation}
\hat{y}_i=\arg\max_{c\in\mathcal{T}} \phi(\hat{x}_i)^{\top}\mathbf{t}_c .
\end{equation}
Thus, a reconstruction is regarded as leaking the original identity if \(\hat{y}_i=y_i\), and as being redirected to another identity if \(\hat{y}_i\neq y_i\). 
Let \(y_i^d\) denote the target decoy identity assigned to sample \(i\). We define
\begin{equation}\begin{aligned}\mathcal{O} &= \{i\in\mathcal{V}\mid \hat{y}_i=y_i\},\\\mathcal{R} &= \{i\in\mathcal{V}\mid \hat{y}_i\neq y_i\},\\\mathcal{D} &= \{i\in\mathcal{V}\mid \hat{y}_i=y_i^d\}.
\end{aligned}
\end{equation}
Here, \(\mathcal{O}\) denotes original-identity leakage, \(\mathcal{R}\) denotes redirection to any non-original identity, and \(\mathcal{D}\subseteq\mathcal{R}\) denotes successful redirection to the target decoy identity. 
The Identity Leakage Ratio (ILR), Identity Redirection Ratio (IRR), and Decoy Hit Ratio (DHR) metrics are defined as
\begin{equation}\mathrm{ILR}=\frac{|\mathcal{O}|}{|\mathcal{V}|},\quad\mathrm{IRR}=\frac{|\mathcal{R}|}{|\mathcal{V}|},\quad\mathrm{DHR}=\frac{|\mathcal{D}|}{|\mathcal{V}|}.
\end{equation}
When \(|\mathcal{V}|=0\), the identity-related metrics are undefined and reported as N/A.
Since existing baselines do not define a target decoy identity, DHR is only reported in the ablation studies.

\begin{table*}[t]
\centering
\renewcommand{\arraystretch}{0.95}
\caption{Comparison of authorized face verification accuracy on standard benchmarks and overall privacy outcome. Bold denotes the best result in each column.} 
\begin{tabular}{lccccccccc}
\toprule
Method & LFW & AgeDB & CFP-FP & CALFW & CPLFW & IJB-B & IJB-C & Privacy Protected &  Imperceptible\\
\midrule
DCTDP            & 99.68 & 97.42 & 95.26 & 95.65 & 90.05 & 93.70 & 95.46 &  $\times$&  $\times$  \\
DuetFace         & \textbf{99.82} & \textbf{97.83} & \textbf{97.87} & \textbf{95.92} & \textbf{92.37} & \textbf{94.00} & \textbf{95.51} &  $\times$&  $\times$  \\
Cloak            & 99.52 & 95.80 & 96.54 & 95.12 & 90.80 & 79.95 & 81.66 &  $\times$&  $\times$  \\
AdvFace          & 99.35 & 94.85 & 93.89 & 94.77 & 89.68 & 75.22 & 82.37 &  $\times$&  $\times$  \\
MinusFace        & 99.67 & 96.13 & 95.03 & 95.38 & 90.18 & 93.37 & 94.70 &  $\checkmark$ &  $\times$\\
FaceObfuscator   & 99.08 & 95.72 & 94.66 & 94.73 & 90.23 & 89.19 & 91.46 & $\checkmark$&  $\times$\\
FracFace         & 99.60 & 96.08 & 94.26 & 95.42 & 88.18 & 85.57 & 87.34 & $\checkmark$&  $\times$\\
\cellcolor{myhighlight}\textbf{Ours} 
& \cellcolor{myhighlight}99.77 
& \cellcolor{myhighlight}97.20 
& \cellcolor{myhighlight}95.54 
& \cellcolor{myhighlight}95.63 
& \cellcolor{myhighlight}91.15 
& \cellcolor{myhighlight}91.10 
& \cellcolor{myhighlight}93.35 
& \cellcolor{myhighlight}$\checkmark$ & \cellcolor{myhighlight}$\checkmark$\\
\bottomrule
\end{tabular}
\end{table*}
\begin{figure*}
    \centering
    \includegraphics[width=\linewidth]{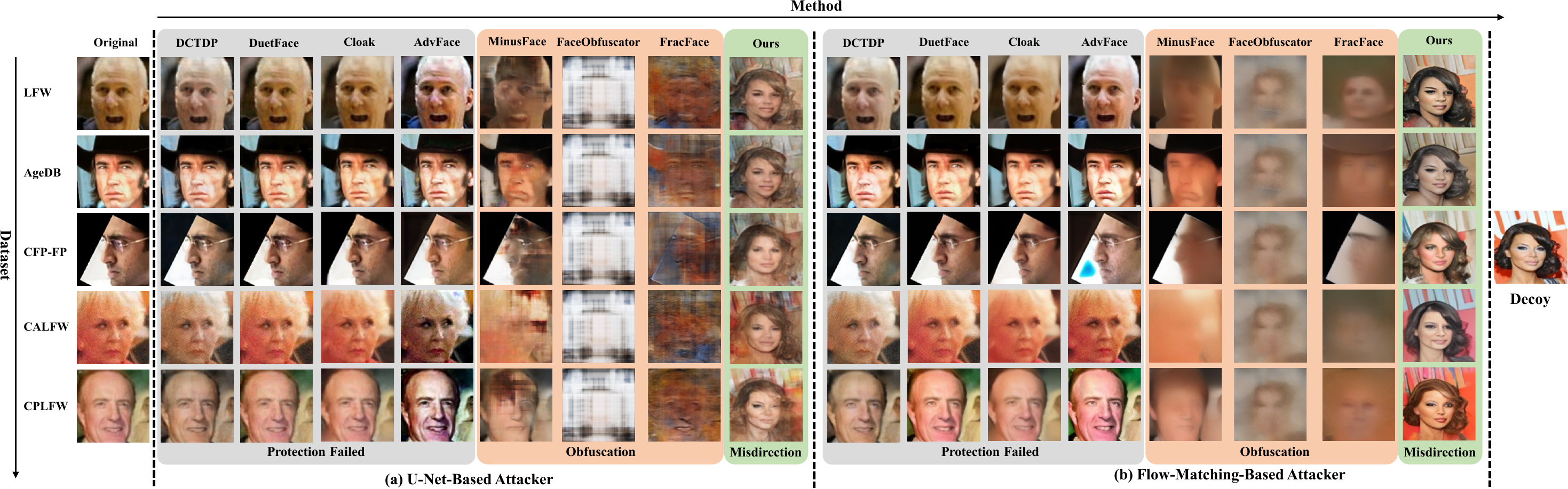}
    \caption{Qualitative reconstruction results under U-Net and Flow-Matching attackers. DecoyFace produces plausible reconstructions that are visually closer to the decoy identity than to the original identity.}
    \label{fig:compare}
\end{figure*}
\subsection{Implementation Details}
We adopt IR-50 \cite{he2016deep} as the backbone and split it into a client-side encoder $F_c$ and a server-side recognizer $F_s$ after the second residual stage, resulting in an intermediate feature of size $64\times 56\times 56$. 

For the RSSD, the complementary projection is constructed based on a reconstruction-sensitive subspace with rank $r=32$. 
The complementary branch is partitioned group-wise with group size 4, yielding $G = 16$ groups in total, among which $K = 3$ groups preserve routed carrier information from the true sample while the remaining groups are replaced using auxiliary complementary decoy features from $D_N$.
The auxiliary complementary bank size is set to $M = 5000$, which is sampled from CelebA \cite{liu2015faceattributes}.
For the coherent-subset policy, $s_c \sim  U[8,12]$, suppressed groups are reassigned to share $L\sim  U[2,4]$ auxiliary sources, while the remaining suppressed groups keep their independent assignments.
To obtain the RSS, we use a U-Net \cite{ronneberger2015u} proxy inversion model trained with $30,000$ iterations on CelebA. 
For transport-time perturbation, we adopt the scale-only variant of the global transformation, where the exponent $k$ is sampled from $[-3, 3]$. 
For server-side canonicalization, we use the group-wise quantile normalizer, with $q = 0.95$ and $\tau_s = 0.8$.

For defender training, we use SGD with momentum 0.9, weight decay $5\times 10^{-4}$. 
The initial learning rate is 0.1, and it decays by a factor of 0.1 at epochs 4, 8, and 12. 
The batch size is 256, and training runs for 15 epochs. 
Following the multi-view formulation in Section IV, we generate two protected views for each training sample, and $\lambda_{\mathrm{cons}}$ is 0.1.
For the attacker, we use a U-Net-based network \cite{ronneberger2015u} and a Flow-Matching-based (FM-based) network \cite{ren2025your}.
Both attackers are trained on CelebA \cite{liu2015faceattributes} for 100,000 iterations.

\begin{table*}[t]
\centering

\label{tab:exp-lfw}
\begin{minipage}[c]{0.52\textwidth}
\centering
\scriptsize
\setlength{\tabcolsep}{2.2pt}
\renewcommand{\arraystretch}{1.3}
\captionof{table}{Privacy utility comparison results on the LFW.}
\resizebox{\linewidth}{!}{
\begin{tabular}{lcccccccc}
\toprule
\multirow{2}{*}{Method} & \multicolumn{4}{c}{U-Net Attacker} & \multicolumn{4}{c}{Flow-Matching Attacker} \\
\cmidrule(lr){2-5} \cmidrule(lr){6-9}
& COS $\downarrow$ & ILR $\downarrow$ & IRR $\uparrow$ & FVR $\uparrow$ & COS $\downarrow$ & ILR $\downarrow$ & IRR $\uparrow$ & FVR $\uparrow$ \\
\midrule
DCTDP & 0.9292 & 100.00 & 0.00 & \textbf{99.92} & 0.7187 & 99.99 & 0.01 & 99.91 \\
DuetFace & 0.9561 & 100.00 & 0.00 & 99.88 & 0.7581 & 100.00 & 0.00 & \textbf{99.93} \\
MinusFace & 0.3500 & 83.61 & 16.39 & 87.02 & 0.1960 & 38.28 & 61.72 & 78.03 \\
FaceObfuscator & \textbf{-0.0051} & N/A & N/A & 0.00 & \textbf{-0.0079} & N/A & N/A & 0.00 \\
FracFace & 0.4424 & 95.96 & 4.04 & 60.16 & 0.2617 & 59.46 & 40.54 & 89.15 \\
Cloak & 0.9420 & 100.00 & 0.00 & 99.90 & 0.6997 & 100.00 & 0.00 & 99.83 \\
AdvFace & 0.9408 & 100.00 & 0.00 & 99.83 & 0.7162 & 100.00 & 0.00 & 99.91 \\
\cellcolor{myhighlight}\textbf{Ours} & \cellcolor{myhighlight}0.0267 & \cellcolor{myhighlight}\textbf{2.93} & \cellcolor{myhighlight}\textbf{97.07} & \cellcolor{myhighlight}99.88 & \cellcolor{myhighlight}0.0137 & \cellcolor{myhighlight}\textbf{0.74} & \cellcolor{myhighlight}\textbf{99.26} & \cellcolor{myhighlight}99.78 \\
\bottomrule
\end{tabular}
}
\end{minipage}
\begin{minipage}[c]{0.44\textwidth}
    \centering
    \includegraphics[width=\linewidth]{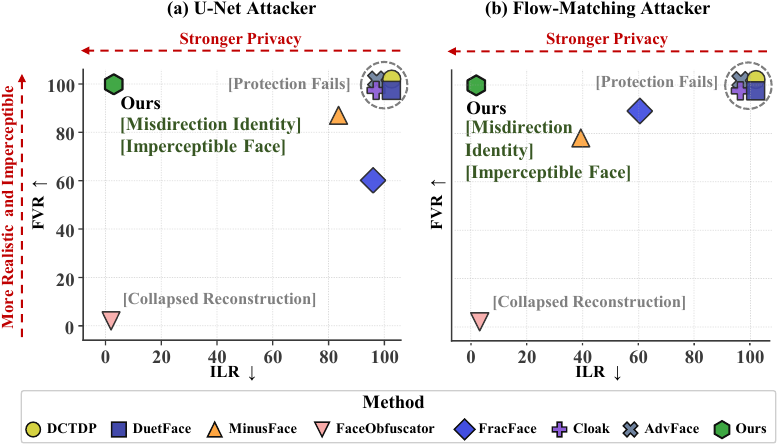}
    \captionof{figure}{Privacy--utility visualization on the LFW. The markers of DCTDP, DuetFace, AdvFace, and Cloak are jittered to avoid overlap.}
\end{minipage}
\end{table*}

\begin{table*}[t]
\centering

\label{tab:exp-agedb}
\begin{minipage}[c]{0.52\textwidth}
\centering
\scriptsize
\setlength{\tabcolsep}{2.2pt}
\renewcommand{\arraystretch}{1.3}
\caption{Privacy utility comparison results on the AgeDB. }
\resizebox{\linewidth}{!}{%
\begin{tabular}{lcccccccc}
\toprule
\multirow{2}{*}{Method} & \multicolumn{4}{c}{U-Net Attacker} & \multicolumn{4}{c}{Flow-Matching Attacker} \\
\cmidrule(lr){2-5} \cmidrule(lr){6-9}
& COS $\downarrow$ & ILR $\downarrow$ & IRR $\uparrow$ & FVR $\uparrow$ & COS $\downarrow$ & ILR $\downarrow$ & IRR $\uparrow$ & FVR $\uparrow$ \\
\midrule
DCTDP & 0.9431 & 99.95 & 0.05 & 98.91 & 0.7085 & 99.86 & 0.14 & 98.98 \\
DuetFace & 0.9641 & 100.00 & 0.00 & 97.55 & 0.7414 & 99.80 & 0.20 & 98.12 \\
MinusFace & 0.4551 & 89.55 & 10.45 & 89.75 & 0.2604 & 53.46 & 46.54 & 83.57 \\
FaceObfuscator & \textbf{0.0010} & N/A & N/A & 0.00 & \textbf{-0.0218} & N/A & N/A & 0.00 \\
FracFace & 0.5427 & 97.51 & 2.49 & 70.73 & 0.3253 & 68.33 & 31.67 & 91.92 \\
Cloak & 0.9307 & 100.00 & 0.00 & 98.22 & 0.6881 & 99.86 & 0.14 & 98.24 \\
AdvFace & 0.9414 & 100.00 & 0.00 & 97.79 & 0.7201 & 99.81 & 0.19 & 98.54 \\
\cellcolor{myhighlight}\textbf{Ours} & \cellcolor{myhighlight}0.0240 & \cellcolor{myhighlight}\textbf{3.38} & \cellcolor{myhighlight}\textbf{96.62} & \cellcolor{myhighlight}\textbf{99.78} & \cellcolor{myhighlight}0.0132 & \cellcolor{myhighlight}\textbf{0.90} & \cellcolor{myhighlight}\textbf{99.10} & \cellcolor{myhighlight}\textbf{99.52} \\
\bottomrule
\end{tabular}
}
\end{minipage}
\begin{minipage}[c]{0.44\textwidth}
    \centering
    \includegraphics[width=\linewidth]{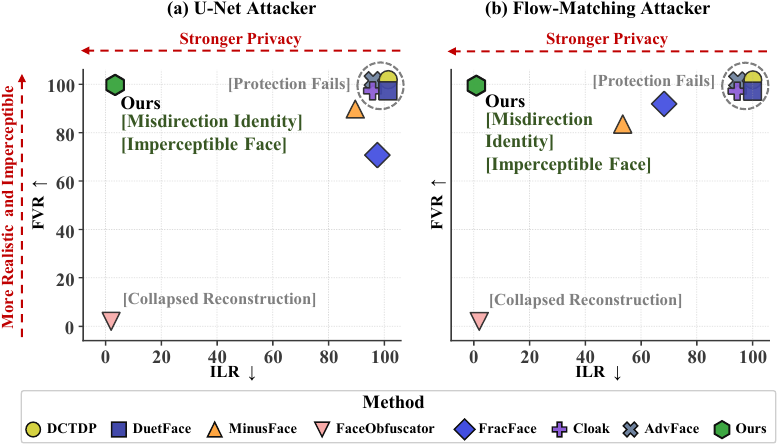}
    \captionof{figure}{Privacy--utility visualization on the AgeDB. The markers of DCTDP, DuetFace, AdvFace, and Cloak are jittered to avoid overlap.}
\end{minipage}
\end{table*}

\begin{table*}[t]
\centering

\label{tab:exp-cfpfp}
\begin{minipage}[c]{0.52\textwidth}
\centering
\scriptsize
\setlength{\tabcolsep}{2.2pt}
\renewcommand{\arraystretch}{1.3}
\caption{Privacy utility comparison results on the CFP-FP.}
\resizebox{\linewidth}{!}{%
\begin{tabular}{lcccccccc}
\toprule
\multirow{2}{*}{Method} & \multicolumn{4}{c}{U-Net Attacker} & \multicolumn{4}{c}{Flow-Matching Attacker} \\
\cmidrule(lr){2-5} \cmidrule(lr){6-9}
& COS $\downarrow$ & ILR $\downarrow$ & IRR $\uparrow$ & FVR $\uparrow$ & COS $\downarrow$ & ILR $\downarrow$ & IRR $\uparrow$ & FVR $\uparrow$ \\
\midrule
DCTDP & 0.9170 & 100.00 & 0.00 & 96.23 & 0.6936 & 100.00 & 0.00 & 96.16 \\
DuetFace & 0.9528 & 100.00 & 0.00 & 95.43 & 0.7391 & 100.00 & 0.00 & 95.89 \\
MinusFace & 0.3584 & 86.95 & 13.05 & 73.91 & 0.2063 & 48.72 & 51.28 & 56.09 \\
FaceObfuscator & \textbf{-0.0182} & N/A & N/A & 0.00 & \textbf{-0.0097} & N/A & N/A & 0.00 \\
FracFace & 0.4662 & 97.93 & 2.07 & 56.34 & 0.2872 & 74.08 & 25.92 & 72.69 \\
Cloak & 0.9139 & 100.00 & 0.00 & 95.53 & 0.6895 & 99.98 & 0.02 & \textbf{99.98} \\
AdvFace & 0.8759 & 100.00 & 0.00 & 89.68 & 0.6510 & 99.76 & 0.24 & 89.82 \\
\cellcolor{myhighlight}\textbf{Ours} & \cellcolor{myhighlight}0.0247 & \cellcolor{myhighlight}\textbf{3.56} & \cellcolor{myhighlight}\textbf{96.44} & \cellcolor{myhighlight}\textbf{99.23} & \cellcolor{myhighlight}0.0177 & \cellcolor{myhighlight}\textbf{1.79} & \cellcolor{myhighlight}\textbf{98.21} & \cellcolor{myhighlight}98.72 \\
\bottomrule
\end{tabular}
}
\end{minipage}
\begin{minipage}[c]{0.44\textwidth}
    \centering
    \includegraphics[width=\linewidth]{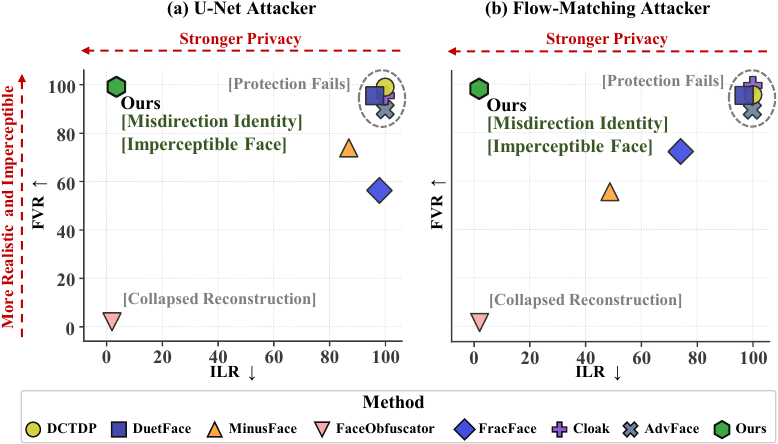}
    
    \captionof{figure}{Privacy--utility visualization on the CFP-FP. The markers of DCTDP, DuetFace, and Cloak are slightly jittered to avoid overlap.}
\end{minipage}
\end{table*}

\begin{table*}[t]
\centering

\label{tab:exp-calfw}
\begin{minipage}[c]{0.52\textwidth}
\centering
\scriptsize
\setlength{\tabcolsep}{2.2pt}
\renewcommand{\arraystretch}{1.3}
\caption{Privacy utility comparison results on the CALFW.}
\resizebox{\linewidth}{!}{%
\begin{tabular}{lcccccccc}
\toprule
\multirow{2}{*}{Method} & \multicolumn{4}{c}{U-Net Attacker} & \multicolumn{4}{c}{Flow-Matching Attacker} \\
\cmidrule(lr){2-5} \cmidrule(lr){6-9}
& COS $\downarrow$ & ILR $\downarrow$ & IRR $\uparrow$ & FVR $\uparrow$ & COS $\downarrow$ & ILR $\downarrow$ & IRR $\uparrow$ & FVR $\uparrow$ \\
\midrule
DCTDP & 0.9400 & 100.00 & 0.00 & 99.58 & 0.7112 & 99.97 & 0.03 & 99.67 \\
DuetFace & 0.9606 & 100.00 & 0.00 & 99.27 & 0.7405 & 99.97 & 0.03 & 99.44 \\
MinusFace & 0.4044 & 91.20 & 8.80 & 90.45 & 0.2273 & 47.48 & 52.52 & 80.88 \\
FaceObfuscator & \textbf{-0.0036} & N/A & N/A & 0.00 & \textbf{-0.0082} & N/A & N/A & 0.00 \\
FracFace & 0.5106 & 98.10 & 1.90 & 67.97 & 0.2923 & 66.33 & 33.67 & 92.01 \\
Cloak & 0.9358 & 100.00 & 0.00 & 99.43 & 0.6801 & 99.97 & 0.03 & \textbf{99.97} \\
AdvFace & 0.9429 & 100.00 & 0.00 & 99.32 & 0.7476 & 99.49 & 0.51 & 99.49 \\
\cellcolor{myhighlight}\textbf{Ours} & \cellcolor{myhighlight}0.0297 & \cellcolor{myhighlight}\textbf{3.86} & \cellcolor{myhighlight}\textbf{96.14} & \cellcolor{myhighlight}\textbf{99.85} & \cellcolor{myhighlight}0.0157 & \cellcolor{myhighlight}\textbf{1.15} & \cellcolor{myhighlight}\textbf{98.85} & \cellcolor{myhighlight}99.64 \\
\bottomrule
\end{tabular}
}
\end{minipage}
\begin{minipage}[c]{0.44\textwidth}
    \centering
    \includegraphics[width=\linewidth]{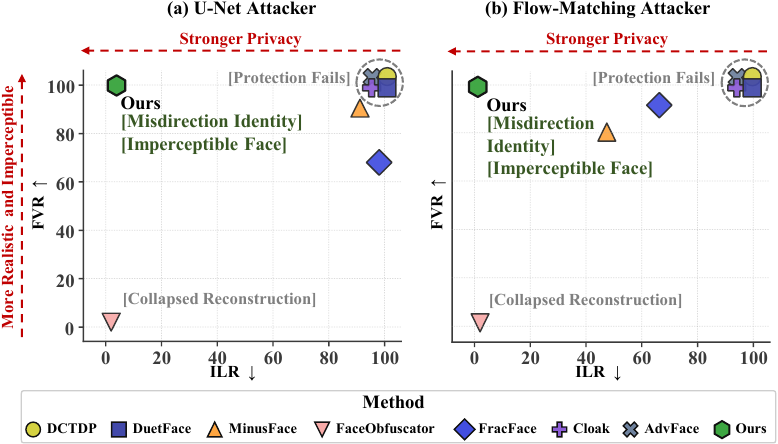}
    \captionof{figure}{Privacy--utility visualization on the CALFW. The markers of DCTDP, DuetFace, AdvFace, and Cloak are jittered to avoid overlap.}
\end{minipage}
\end{table*}

\begin{table*}[t]
\centering

\label{tab:exp-cplfw}
\begin{minipage}[c]{0.52\textwidth}
\centering
\scriptsize
\setlength{\tabcolsep}{2.2pt}
\renewcommand{\arraystretch}{1.3}
\caption{Privacy utility comparison results on the CPLFW. }
\resizebox{\linewidth}{!}{%
\begin{tabular}{lcccccccc}
\toprule
\multirow{2}{*}{Method} & \multicolumn{4}{c}{U-Net Attacker} & \multicolumn{4}{c}{Flow-Matching Attacker} \\
\cmidrule(lr){2-5} \cmidrule(lr){6-9}
& COS $\downarrow$ & ILR $\downarrow$ & IRR $\uparrow$ & FVR $\uparrow$ & COS $\downarrow$ & ILR $\downarrow$ & IRR $\uparrow$ & FVR $\uparrow$ \\
\midrule
DCTDP & 0.9021 & 100.00 & 0.00 & 93.31 & 0.7221 & 99.98 & 0.02 & 93.24 \\
DuetFace & 0.9502 & 100.00 & 0.00 & 92.27 & 0.7858 & 100.00 & 0.00 & 92.24 \\
MinusFace & 0.3502 & 80.52 & 19.48 & 66.13 & 0.1985 & 40.54 & 59.46 & 51.08 \\
FaceObfuscator & \textbf{-0.0118} & N/A & N/A & 0.00 & \textbf{-0.0070} & N/A & N/A & 0.00 \\
FracFace & 0.4072 & 95.40 & 4.60 & 42.73 & 0.2630 & 63.84 & 36.16 & 68.35 \\
Cloak & 0.9223 & 100.00 & 0.00 & 92.46 & 0.7146 & 100.00 & 0.00 & 90.81 \\
AdvFace & 0.9302 & 100.00 & 0.00 & 92.19 & 0.7476 & 99.99 & 0.01 & 92.57 \\
\cellcolor{myhighlight}\textbf{Ours} & \cellcolor{myhighlight}0.0301 & \cellcolor{myhighlight}\textbf{2.63} & \cellcolor{myhighlight}\textbf{97.37} & \cellcolor{myhighlight}\textbf{99.45} & \cellcolor{myhighlight}0.0216 & \cellcolor{myhighlight}\textbf{1.59} & \cellcolor{myhighlight}\textbf{98.41} & \cellcolor{myhighlight}\textbf{98.98} \\
\bottomrule
\end{tabular}
}
\end{minipage}
\begin{minipage}[c]{0.44\textwidth}
    \centering
    \includegraphics[width=\linewidth]{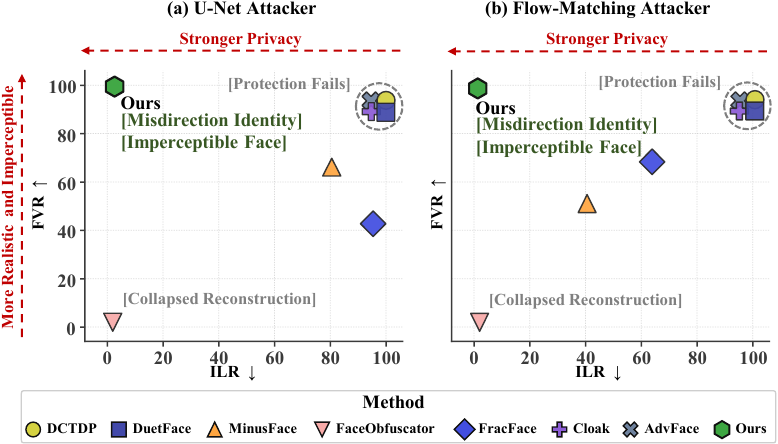}
        \captionof{figure}{Privacy--utility visualization on the CPLFW. The markers of DCTDP, DuetFace, AdvFace, and Cloak are jittered to avoid overlap.}
\end{minipage}
\end{table*}

We compare the proposed method with representative PPFR methods\footnote{As official implementations of AdvFace, FaceObfuscator, and FracFace are unavailable, we reproduce them based on the details reported in their papers. For MinusFace, we report the IJB-B and IJB-C results from the original paper.}, including DCTDP \cite{ji2022privacy}, DuetFace \cite{mi2022duetface}, MinusFace \cite{mi2024privacy}, FaceObfuscator \cite{jin2024faceobfuscator}, FracFace \cite{daifracface}, Cloak \cite{mireshghallah2021not}, and AdvFace \cite{wang2023privacy}. 
These baselines cover several major design strategies, including frequency-domain collaboration, adversarial feature obfuscation, and inversion-resistant feature learning. All experiments were conducted on NVIDIA A100 GPUs.
\subsection{Recognition Accuracy}
Table II reports the authorized recognition performance on standard face verification benchmarks. 
Overall, DecoyFace preserves competitive recognition performance across all datasets.
Although DuetFace and MinusFace achieve higher Acc on some benchmarks, DecoyFace provides a more favorable privacy–utility trade-off by maintaining competitive authorized recognition performance while achieving identity misdirection rather than identity leakage or perceptible obfuscation.
Specifically, DecoyFace achieves $99.77\%$ on LFW, $97.20\%$ on AgeDB, $95.54\%$ on CFP-FP, $95.63\%$ on CALFW, and $91.15\%$ on CPLFW.
On the more challenging template-based benchmarks, it obtains $91.10\%$ TAR@FAR=$10^{-4}$ on IJB-B and $93.35\%$ on IJB-C.
In particular, it outperforms FaceObfuscator, FracFace, Cloak, and AdvFace on most verification benchmarks.
These results suggest that the proposed DecoyFace preserves most of the identity-discriminative information required for authorized verification, despite deliberately perturbing the transmitted representation to protect privacy.
It is worth noting that DecoyFace does not pursue recognition accuracy alone, but a better privacy--utility trade-off, as shown in the following section.
Some methods improve utility but still leak the original identity under inversion.
By contrast, DecoyFace retains competitive verification utility while explicitly targeting identity redirection, which better aligns with the practical objective of PPFR.

\subsection{Evaluation of Privacy Protection}
\subsubsection{Privacy Against Malicious Attackers - \textbf{Any Way}}
Tables III-VII and Figs. 5-10 report reconstruction privacy against malicious attackers under a U-Net-based attacker and a FM-based attacker. 
Fig. 5 shows qualitative reconstruction examples under both U-Net-based and FM-based attackers. Prior methods either preserve the original identity in visually plausible reconstructions or produce severely degraded outputs that expose the presence of protection. In contrast, DecoyFace maintains facial plausibility while visually redirecting the reconstruction toward the decoy identity, which is consistent with its low ILR and high FVR.
It can be observed that prior methods mainly fall into two patterns, whereas DecoyFace exhibits a markedly different privacy profile.
\ding{172} \textbf{Privacy Leakage Defense}: The first group, including DCTDP, DuetFace, Cloak, and AdvFace, preserves highly face-valid reconstructions, with FVR remaining close to $100\%$.
However, these methods still exhibit severe identity leakage: their ILR remains close to $100\%$, whereas IRR is close to $0\%$.
This means that although the reconstructed faces remain visually plausible, they are still recognized as the original identity in almost all cases.
Therefore, preserving visual realism while slightly degrading reconstruction fidelity is insufficient for preventing identity leakage.
\ding{173} \textbf{Effective-but-obvious Defenses}: The second group corresponds to stronger suppression-oriented defenses.
FaceObfuscator collapses the reconstruction, leading FVR to 0\%, so identity metrics become undefined rather than indicating successful misdirection.
FracFace and MinusFace partially reduce privacy leakage, but still fail to maintain high face validity and strong identity redirection simultaneously.

These results suggest that prior methods primarily trade off between two unsatisfactory extremes: either the reconstructed result still reveals the original identity, or the reconstruction quality deteriorates to the point that the defense becomes obvious.
In contrast, DecoyFace achieves a different privacy profile.
Under the malicious-attacker setting, DecoyFace maintains consistently high face validity, with FVR ranging from $98.72\%$ to $99.88\%$, indicating that the reconstructed outputs are still plausible faces.
At the same time, the reconstructed results exhibit very weak consistency with the original identity, as reflected by the low COS across all benchmarks.
More importantly, DecoyFace reduces ILR to $2.63\%$--$3.86\%$ under the U-Net attacker and $0.74\%$--$1.79\%$ under the FM attacker.
This means that in most cases, the attacker still obtains a valid face, but it is recognized as a non-original identity rather than the true subject.
This observation is consistent with the decoy-oriented objective of DecoyFace: the defense does not merely make reconstruction fail, but instead redirects privacy leakage toward a realistic yet incorrect identity.
\begin{table}[t]
\centering
\caption{Privacy under HBC Server-Side Reconstruction.}
\renewcommand{\arraystretch}{0.9}
\begin{tabular}{llcccc}
\toprule
Dataset&Method & COS$\downarrow$ & ILR$\downarrow$&IRR$\uparrow$ &FVR$\uparrow$ \\
\midrule
\multirow{2}{*}{LFW}&FaceObfuscator&0.4898&88.33 &11.67&89.45\\
&\cellcolor{myhighlight}\textbf{Ours}    & \cellcolor{myhighlight}\textbf{0.0059} & \cellcolor{myhighlight}\textbf{0.04} &\cellcolor{myhighlight}\textbf{99.96} &\cellcolor{myhighlight}\textbf{99.98}  \\
\midrule
\multirow{2}{*}{AgeDB}&FaceObfuscator&0.5185 &85.36&14.64&90.57\\
&\cellcolor{myhighlight}\textbf{Ours}    
&\cellcolor{myhighlight} \textbf{0.0007}
&\cellcolor{myhighlight}\textbf{0.08} & \cellcolor{myhighlight}\textbf{99.92} &\cellcolor{myhighlight}\textbf{99.99} \\
\midrule
\multirow{2}{*}{CFP-FP}&FaceObfuscator&0.4228 &85.36&14.64&78.89\\
&\cellcolor{myhighlight}\textbf{Ours}    & \cellcolor{myhighlight}\textbf{0.0046} & \cellcolor{myhighlight}\textbf{0.34 }&\cellcolor{myhighlight}\textbf{99.66} &\cellcolor{myhighlight}\textbf{99.93}  \\
\midrule
\multirow{2}{*}{CALFW}&FaceObfuscator&0.5038&89.41 &10.59&89.25\\
&\cellcolor{myhighlight}\textbf{Ours}    & \cellcolor{myhighlight}\textbf{0.0067} & \cellcolor{myhighlight}\textbf{0.07} &\cellcolor{myhighlight}\textbf{99.93} &\cellcolor{myhighlight}\textbf{100.00}  \\
\midrule
\multirow{2}{*}{CPLFW}&FaceObfuscator&0.4134&83.57 &16.43&70.64\\
&\cellcolor{myhighlight}\textbf{Ours}    & \cellcolor{myhighlight}\textbf{0.0094} & \cellcolor{myhighlight}\textbf{0.08 }&\cellcolor{myhighlight}\textbf{99.92 }&\cellcolor{myhighlight}\textbf{99.94}  \\
\bottomrule
\end{tabular}
\end{table}
\subsubsection{Privacy Against HBC Servers - \textbf{Any Where}}
Different from the malicious attacker that reconstructs from the transmitted protected feature $z_t$, the HBC server trains its reconstruction model directly on the server-side canonicalized feature $z_{can}$. 
We assume that the HBC server knows the recovery pipeline, including the projection $P_N$, quantile calibration, and the RSS anchor $\mu_{\mathrm{rec}}$. This setting is more challenging because $z_{can}$ is the representation used for authorized recognition and therefore inevitably retains certain identity-discriminative evidence.
Since most existing baselines do not introduce an additional server-side canonicalization module comparable to $z_{can}$, we include FaceObfuscator as a representative baseline for the HBC comparison, as shown in Table VIII.
FaceObfuscator still suffers from substantial original-identity leakage, with ILR ranging from 83.57\% to 89.41\%, whereas DecoyFace reduces ILR to 0.04\%–0.34\% while keeping FVR between 99.93\% and 100.00\%.
Existing PPFR methods mainly protect the transmitted representation, but do not explicitly prevent a server-side recognition feature from being exploited for inversion. 
In contrast, DecoyFace does not recover the original intermediate feature on the server side. The RSS component is suppressed by $P_N$, while the complementary carrier has already been sparsified and mixed with auxiliary identities through DGCM. 
Quantile calibration and RSS anchor $\mu_{rec}$ restore a stable distribution for recognition, but do not restore a clean complementary branch of the true sample. 
Therefore, $z_{can}$ remains sufficiently discriminative for authorized verification, yet lacks the visual and identity-complete evidence required for faithful reconstruction by the HBC server.

\begin{table}[t]
\caption{Component ablations of DecoyFace.}
\label{tab:client_design_ablation}
\centering
\setlength{\tabcolsep}{6pt}
\renewcommand{\arraystretch}{0.9}
\begin{tabular}{l c c c c}
\toprule
Setting
& Acc $\uparrow$
& ILR $\downarrow$
& DHR $\uparrow$
& FVR $\uparrow$ \\
\midrule
w/o RSSD & 61.37 & 94.32 & 0.40 & 92.47 \\
w/o DGCM & 91.97 & 100.00 & 0.00  & 92.83 \\
w/o ACM & 66.02 & 2.63 &30.19& 99.45\\
\midrule
w/o CSG & 90.12 & 53.24 & 7.12& 95.57 \\
w/o IN & 89.20 & 53.29 & 6.40 & 95.45 \\
w/o Coherent & 90.11&86.69 & 0.07 & 94.72 \\
w/o Scale & 90.68 &72.89 & 2.88 &94.20 \\
\midrule
w/o Calib & 81.38 & 2.63 &30.19& 99.45 \\
w/o $\mu_{\mathrm{rec}}$ & 83.68 & 2.63 &30.19& 99.45 \\
w/o $\mathcal{L}_{cons}$ & 88.77 & 2.63 &30.19& 99.45\\
\midrule
\cellcolor{myhighlight}\textbf{Full} & \cellcolor{myhighlight}91.15 & \cellcolor{myhighlight}2.63&\cellcolor{myhighlight}30.19&\cellcolor{myhighlight}99.45 \\
\bottomrule
\end{tabular}
\end{table}

\begin{table}[t]
\caption{EFFECT OF GROUP NUMBER G AND PRESERVED TRUE-SAMPLE GROUPS K IN DGCM.}
\label{tab:gk_ablation}
\centering
\renewcommand{\arraystretch}{0.9}
\setlength{\tabcolsep}{6pt}
\begin{tabular}{c cc c cc}
\toprule
G &K& Acc $\uparrow$
& ILR $\downarrow$
& DHR $\uparrow$
& FVR $\uparrow$ \\
\midrule
16 & 2 &80.98& 0.12 & 26.78 & 99.88 \\
\cellcolor{myhighlight}\textbf{16}& \cellcolor{myhighlight}\textbf{3} & \cellcolor{myhighlight}91.15 &  \cellcolor{myhighlight}2.63&\cellcolor{myhighlight}30.19&\cellcolor{myhighlight}99.45  \\
16 & 4 &91.17&83.48 & 4.92 & 94.20 \\
32 & 6 & 88.65& 63.65 &5.08 & 94.88 \\
\bottomrule
\end{tabular}
\end{table}

\subsection{Ablation Studies}
We conduct ablation studies on the CPLFW dataset to analyze the contribution of each component in DecoyFace with the U-Net attacker. 

\subsubsection{Effect of core components}
Table IX summarizes the effect of the major components in the proposed framework.
Removing RSSD severely degrades both utility and privacy, reducing the Acc to 61.37\% on CPLFW, while increasing ILR to 94.32\% and reducing DHR to 0.40\%.
This indicates that RSSD is essential for separating reconstruction-dominant directions from the complementary carrier used for recognition. 
Removing DGCM leads to a different failure mode.
Although the recognition Acc remains relatively high, the privacy protection completely fails, as ILR reaches 100.00\% and DHR drops to 0.00\%.
This result confirms that explicitly constructing the decoy carrier is necessary for steering unauthorized reconstruction away from the original identity.
Removing ACM mainly damages authorized utility, with the Acc dropping from 91.15\% to 66.02\%.
This confirms that the matched server-side ACM is critical for suppressing the decoy-dominant component and restoring a recognition-friendly representation.

\subsubsection{Effect of client-side DGCM}
We further analyze the detailed design choices in DGCM.
Removing the CSG, IN, coherent-subset group mixing policy, and random scaling consistently weakens the identity redirection capability.
Without CSG, ILR increases to 53.24\%, and DHR decreases to 7.12\%, suggesting that randomized grouping helps disrupt residual true-identity traces in the complementary branch.
Without IN, ILR rises to 53.29\%, indicating that statistical alignment is important for stabilizing the mixed representation.
The coherent mixing strategy is particularly important. 
Removing it increases ILR to 86.69\% and reduces DHR to 0.07\%, which shows that independently replaced groups introduce fragmented complementary evidence and weaken decoy-consistent reconstruction.
Removing the random scale also increases ILR to 72.89\% and reduces DHR to 2.88\%, verifying that sample-wise magnitude variation provides an additional barrier against feature inversion.
Notably, all variants maintain high FVR, which indicates that the degradation is not caused by reconstruction collapse.
Instead, these components mainly determine whether a valid reconstructed face is attributed to the original identity or redirected to the decoy identity.

\subsubsection{Effect of server-side ACM design}
The server-side components mainly affect authorized recognition rather than attacker-side privacy. 
The attacker-side privacy metrics remain unchanged because the malicious attacker reconstructs from the transmitted protected feature before server-side processing.
Removing quantile calibration reduces the Acc to 81.38\%, while the privacy metrics remain unchanged.
A similar trend is observed when removing the reconstruction-sensitive anchor $\mu_{\mathrm{rec}}$, where the Acc drops to 83.68\%.
Removing the consistency loss decreases the Acc to 88.77\%, indicating that multi-view consistency helps the server learn embeddings that are robust to stochastic client-side protection.
Overall, these results show that quantile calibration, the reconstruction-sensitive anchor, and the consistency constraint mainly improve the stability of authorized recognition.

\subsubsection{Effect of Group Number and Preserved Groups}
Table~\ref{tab:gk_ablation} studies how the group number $G$ and the number of preserved groups $K$ affect recognition and privacy.
With $G=16$ and $K=2$, the model achieves a very low ILR of 0.12\%, but the Acc drops to 80.98\%, indicating that preserving too few true-sample groups can weaken authorized verification.
Increasing $K$ to 3 improves the Acc to 91.15\%, while keeping ILR low at 2.63\% and maintaining a high DHR of 30.19\%. 
However, further increasing $K$ to 4 causes ILR to rise sharply to 83.48\%, showing that the attacker can exploit excessive true-identity evidence.
When the group number is increased to $G=32$ with $K=6$, ILR remains high at 63.65\%, and Acc drops to 88.65\%, suggesting that overly fine-grained grouping may fragment the mixed representation and weaken coherent decoy guidance.
Therefore, we adopt $G=16$ and $K=3$ as the default configuration.
\begin{figure}
    \centering
    \includegraphics[width=0.9\linewidth]{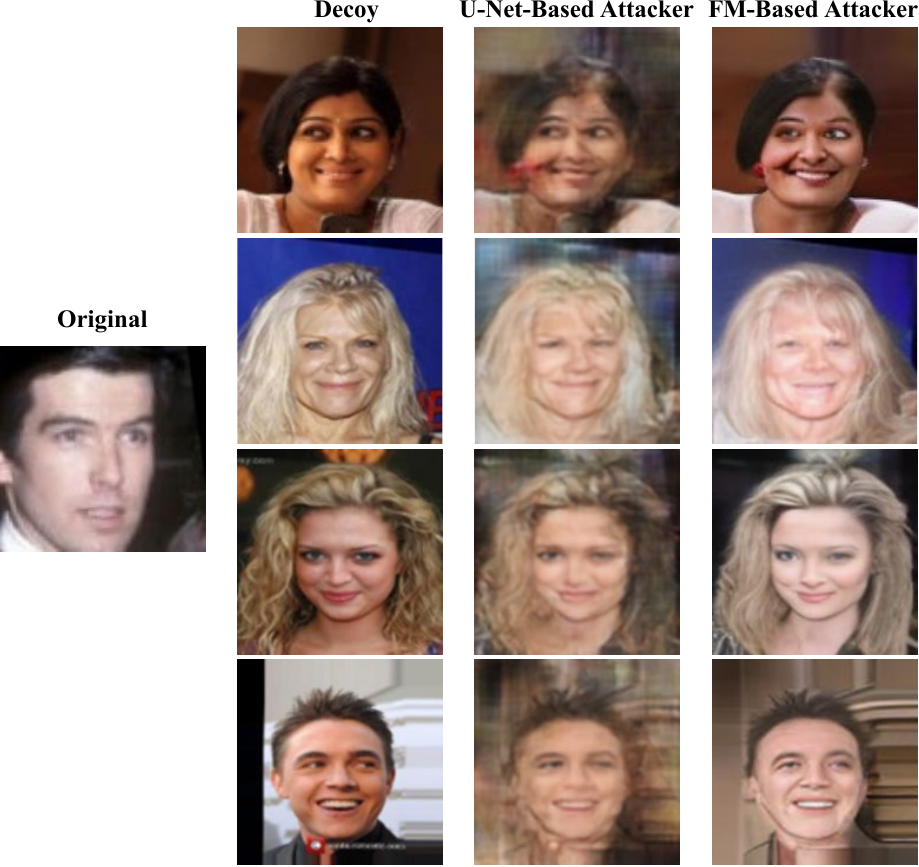}
    \caption{Qualitative results with different decoy identities. The original face is fixed, while each row uses a different target decoy.}
    \label{fig:decoy}
    \vspace{-5mm}
\end{figure}
\subsubsection{Effect of Different Decoy -\textbf{ Any Who}}
To further illustrate the controllability of decoy-oriented misdirection, Fig. \ref{fig:decoy} presents qualitative examples with different decoys. Given the same original face, changing the assigned decoy leads to different unauthorized reconstructions under both U-Net-based and FM-based attackers. The reconstructed faces remain visually plausible, but are shifted away from the original identity and toward the selected decoys.
These results confirm that DecoyFace can achieve controllable identity misdirection.

\section{Conclusion}
We propose DecoyFace, a decoy-oriented privacy-preserving framework for split face recognition. 
Unlike existing PPFR methods that mainly suppress reconstruction quality, DecoyFace aims to misdirect unauthorized reconstruction toward a plausible but incorrect identity while preserving recognition utility for the authorized server. The framework is built upon RSSD, client-side DGCM, and server-side ACM. 
This design injects decoy information into reconstruction-sensitive directions, limits true-identity exposure in the complementary carrier, and enables the server to recover a canonicalized representation for verification.

Experiments demonstrate that DecoyFace preserves competitive recognition accuracy while significantly reducing true-identity leakage. 
Under both malicious attacker-side inversion and HBC server-side reconstruction, DecoyFace maintains high face validity and redirects most unauthorized reconstructions away from the original identity. 
These results show that privacy protection in split face recognition need not rely solely on perceptible obfuscation or reconstruction collapse. 
Instead, decoy-based misdirection provides a stealthier and more controllable privacy-preserving direction.

\footnotesize
\bibliographystyle{unsrt}
\bibliography{ref}

\end{document}